\documentclass[conference]{IEEEtran}
\IEEEoverridecommandlockouts
\usepackage{amsmath,amssymb,amsfonts}
\usepackage{algorithmic}
\usepackage{graphicx}
\usepackage{textcomp}
\usepackage{xcolor}
\usepackage{cleveref}
\usepackage[numbers]{natbib}
\def\BibTeX{{\rm B\kern-.05em{\sc i\kern-.025em b}\kern-.08em
        
\usepackage{natbib}
    T\kern-.1667em\lower.7ex\hbox{E}\kern-.125emX}}

\begin{document}

\title{
    Multi-granulariy Time-based Transformer for Knowledge Tracing
}

\author{\IEEEauthorblockN{Tong Zhou}
\IEEEauthorblockA{\textit{Department of Computer Science} \\
\textit{Johns Hopkins University}\\
Baltimore, United States \\
tzhou11@jhu.edu}
}

\maketitle

\begin{abstract}
    In this paper, we present a transformer architecture for predicting student performance on standardized tests. Specifically, we leverage students' historical data, including their past test scores, study habits, and other relevant information, to create a personalized model for each student. We then use these models to predict their future performance on a given test. Applying this model to the RIIID dataset, we demonstrate that using multiple granularities for temporal features as the decoder input significantly improve model performance. Our results also show the effectiveness of our approach, with substantial improvements over the LightGBM method. Our work contributes to the growing field of AI in education, providing a scalable and accurate tool for predicting student outcomes.
\end{abstract}


\section{Introduction}
Knowledge tracing is an important field of research in educational data mining, as it can help to improve the effectiveness and efficiency of learning. 
The first application is personalized learning. \cite{mahajan2020} surveys that by tracking the histories of individual students over time, instructors can tailor instructional materials to meet specific needs of each student. The second application is early intervention. \cite{corbett1994} recognizes that knowledge tracing enables instructors to intervene early and offer extra assistance or resources to students who need it in order to succeed by identifying those who are having difficulty with specific concepts or skills. Knowledge tracing can also facilitate adaptive learning, which can change the level of difficulty and pace of instruction based on the understanding and performance of individual students. \cite{heffernan2014} shows that learning and engagement can be enhanced since students are appropriately challenged can can avoid frustration or boredom. 

Recent advances in artificial intelligence (AI) have shown great promise for improving educational outcomes. In particular, deep learning models have been developed to predict student performance on standardized tests, such as the SAT and TOEIC, based on a variety of factors, including their previous academic records, socio-economic background, and personal characteristics. One of the key challenges in building such models is capturing the dynamic and complex nature of student behavior, which can vary widely over time and across individuals.

In this paper, we propose a Transformer that leverages users' histories for educational performance prediction. The Transformer is a state-of-the-art neural network architecture that has achieved impressive results in various natural language processing tasks, such as machine translation, question answering, and text generation. Our approach extends the Transformer to model student performance by incorporating their past academic records, study habits, and other relevant information.

To evaluate the effectiveness of our approach, we conducted experiments on real-world educational datasets, including the Kaggle Riiid AIEd Challenge dataset. We demonstrate that converting temporal features into multiple categorical features with different granularities can greatly improve model's performance. The results even show that the lecture information is irrelevant in the present of the multi-granularity temporal features. Our results also demonstrate that our model outperforms traditional LightGBM, achieving state-of-the-art performance in predicting student performance on standardized tests. Moreover, we show that our model can be used to provide personalized recommendations to students based on their historical data, enabling them to improve their academic performance.

\section{Related Work}
Works of knowledge tracing models mainly follow two different approaches: BKT (Bayesian Knowledge Tracing) and DKT (Deep Knowledge Tracing). BKT is a probabilistic model where student knowledge is modeled as a latent variable and other observed context information and learning performance are used to identify the latent structure represented by a Hidden Markov Model. Hidden Markov Model (HMM) is a statistical model used to analyze sequential data. HMMs are made up of a number of observable states, which reflect the observed data, and a number of hidden states, which represent the unobserved variables underlying the data. The Baum-Welch algorithm is used to estimate the transition probabilities between hidden states and the emission probabilities from hidden states to observable states using the training data. HMMs have been widely used in speech recognition \cite{bahl1986maximum}, network analysis \cite{zhuo2017website}, and even social sciences \cite{10206080}. HMM's applications in knowledge tracing can be found in  \cite{corbett1994,kasurinen2009estimating,pardos2010navigating}. BKT, however, has some innate limitations, such as its implausible assumption of independence between skills, and its inability to address correlations between different skills or knowledge components.  

The Deep Knowledge Tracing (DKT) has received increasing attentions.  DKT relies on a recurrent neutral network (RNN) architecture, using a Long Short-Term Memory (LSTM) network to take a sequence of student responses and other context information as input. By its design, DKT is able to handle correlations between different knowledge components and capture complex interactions between these components over time \cite{jeon2021last}.

Though using LightGBM to tackle sequence prediction problems has been popular (\cite{zhou2023improved}) , using Transformer architecture has gained more popularity in recent years, as its self attention mechanism has demonstrated great effectiveness for sequential prediction tasks (\cite{tran2021riiid,oya2021lstm,zhang2021muse}). Three notable papers based on Transformer are SAKT \cite{pandey2019self}, AKT \cite{ghosh2020context}, SAINT \cite{choi2020ednet} and SAINT+ \cite{shin2021saint+}. SAINT+ is mostly related to our paper. We present a slightly different neural network architecture and a notable improvement in feature engineering for temporal features. 

\section{Methods}

\begin{figure}
    \centering
    \includegraphics[scale=.45]{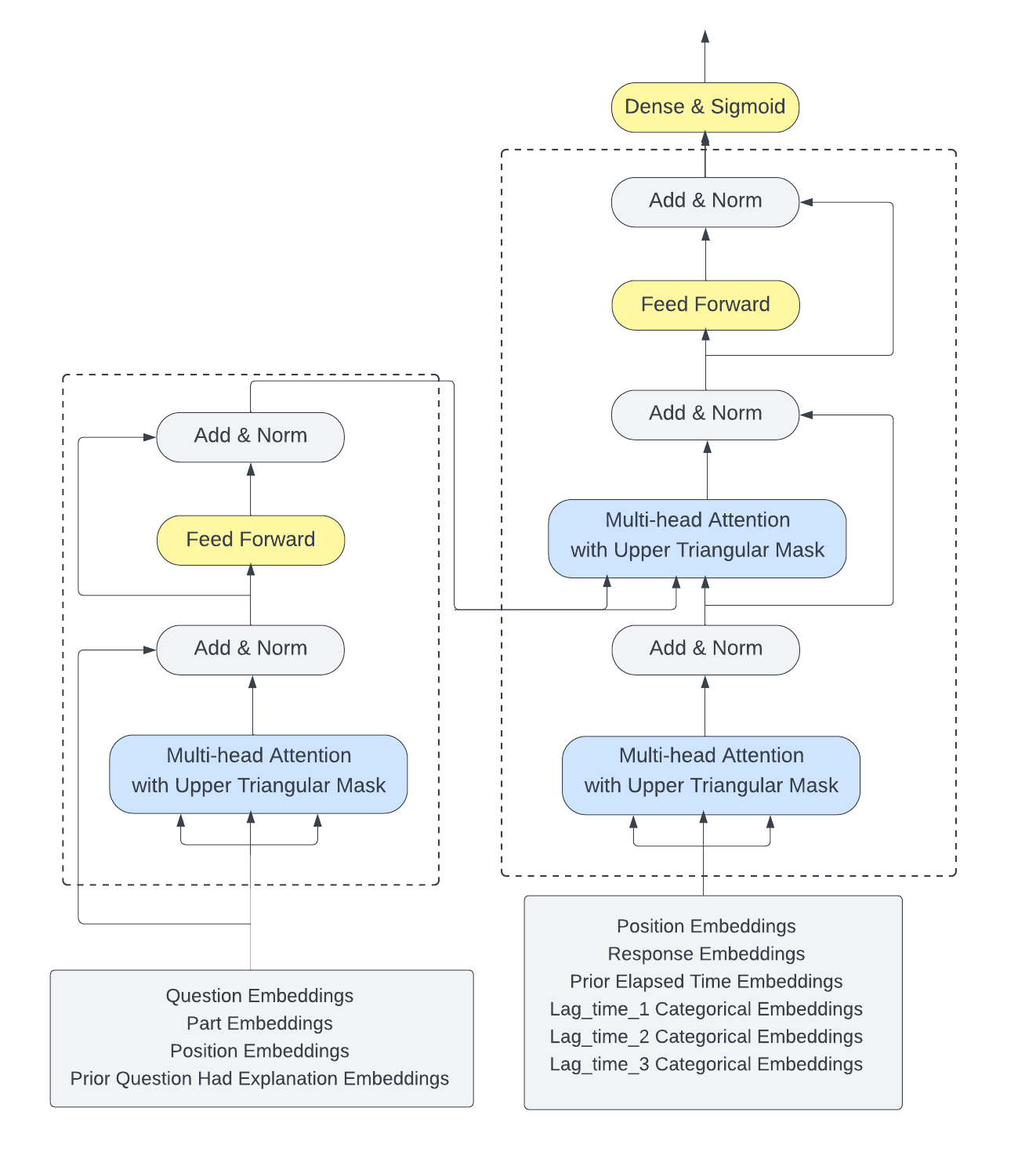}
    \caption{Model Architecture}
    \label{fig}
\end{figure}

We present our Transformer architecture in \Cref{fig}. The encoder in our model consists of four components: question embeddings, part embeddings, position embeddings and prior-question-had-explanation embeddings. The decoder in our model consists of 6 components: position embeddings, response embeddings, prior-elapsed-time embeddings, lag-time-1 categorical embeddings, lag-time-2 categorical embeddings, lag-time-3 categorical embeddings.One of the notable features of our Transformer architecture is that we included three separate embedding layers for representing lag time in three different temporal granularities: by seconds, by minutes, and by days.

This architecture captures multiple types of information. The encoder captures information about the questions and parts of the questions, as well as the presence or absence of explanations. This allows the model to capture multiple types of information that can affect student learning, and to learn meaningful representations of these factors.

It also incorporates temporal information. The encoder uses position embedding to encode the position of each token within the sequence of interactions between the student and the instruction materials. This allows the model to capture the temporal dynamics of the data and learn representations that can account for changes in student learning over time.

It can model complex interactions: The use of multiple embedding layers in both the encoder and decoder components of the model allows for the modeling of complex interactions between different factors that can affect student learning. For example, the model can learn to capture the relationship between a student's prior response and their subsequent performance on a related question, or the interaction between elapsed time and lag time in predicting student learning outcomes.

\subsection{Data}
We use the dataset provided by Kaggle from the "Riiid! Answer Correctness Prediction" competition\footnote{https://www.kaggle.com/c/riiid-test-answer-prediction/}, which challenges participants to build machine learning models that can predict students' responses to questions on an educational platform. 

The dataset contains over 100 million rows of data from student interactions with the platform, including information on the questions, answers, and explanations provided, as well as metadata such as the time elapsed since the previous interaction and the student's performance history. The goal is to predict whether or not a student will correctly answer the next question in a sequence, based on their past interactions with the platform. 
\subsection{Transformer}
Our knowledge tracing model is based Transformer \cite{vaswani2017attention} architecture which includes an encoder and a decoder. A student's interactions with the online learning system are modeled as a sequence of events, where each event corresponds to a particular action taken by the student. These actions may include answering a question, skipping a question, or requesting help, among others. To convert this sequence of events into a format that can be used by the Transformer, the data is typically preprocessed and encoded in two separate steps: one for the encoder and one for the decoder. 

Our paper makes use of upper triangular masks as the multi-head attention (and self-attention). This setting in both the encoder and decoder is only allowed to attend to positions that have already been processed. In other words, each token can only attend to previous tokens in the sequence, but not to tokens that come after it. In order to avoid the model from ``cheating" by taking information from the future, it first makes sure that the model only uses data that is available at each stage of the sequence. 

\subsection{Encoder}
\begin{enumerate}
    \item \textbf{Question embeddings}: this feature encodes question ID. We used an embedding layer of $128$ dimensions. The question embedding layer is intended to help the neural network identify the connections between various abilities or concepts and their significance to the student's overall learning development. The network may learn to spot links between question difficulty and student performance by modeling the questions and tasks using an embedding layer. 
    \item \textbf{Part embeddings}: this feature encodes different sections of test in dataset. This embedding layer is to help the neural network understand the connections between different parts of the questions and their significance to the student's overall learning progress. The network may learn to capture the connections between category difficulty and student perfomance by encoding the categories using an embedding layer. 
    \item \textbf{Position embeddings}: this feature serves to encode the relative position of each token in the input sequence, since the self-attention mechanism treats all tokens equally regardless of their position in the sequence.
    \item \textbf{Prior-question-had-explanation embeddings}: this feature encodes whether or not a user saw the explanation of a question after answering it, with a vocabulary size of $3$. This embedding layer is intended to help the neural network identify the connection between the user's attempt to see an explanation and their overall learning outcomes.
\end{enumerate}

\subsection{Decoder}

\begin{enumerate}
    \item \textbf{Position embeddings}: this feature serves to encode the relative position of each token in the input sequence, since the self-attention mechanism treats all tokens equally regardless of their position in the sequence. 
    \item \textbf{response embeddings}: this feature encodes if a user's answer if correct or not, with a vocabulary size of 3. The goal of the response embedding layer is to help the neural network understand the connection between the accuracy of the student's response and their overall progress of learning. The network can learn to distinguish between right and wrong responses can can record patterns and links between response accuracy and student performance by modelling the response using an embedding layer. This can increase the network's predictive accuracy and make it easier to spot areas where a student may be struggling or excelling.
    \item \textbf{prior-elapsed-time embeddings}: this feature encodes the amount of time that a user used to answer the question, with a vocabulary size of $301$. The goal of the elapsed time embedding is to assist the neural network in learning meaningful representations of the temporal information and in capturing the amount of time that has passed between the current taks or question ant the previous one. The network may learn to distinguish between various time periods and capture patterns and correlations between elapsed time and student performance by expressing elapsed time using an embedding layer. This can increase the network's prediction accuracy and make it easier to spot long-term trends in student learning. 
    \item \textbf{lag-time-1 categorical embeddings}: this feature encodes lag time in seconds, with a vocabulary size of $301$.
    \item \textbf{lag-time-2 categorical embeddings}: this feature encodes lag time for representing lag time in minutes, with a vocabulary size of $1441$.
    \item \textbf{lag-time-3 categorical embedding}: this feature encodes lag time for representing lag time in days, with a vocabulary size of $366$. 
\end{enumerate}
\subsection{Training}
The predicted probability of the user's response being correct at each position is obtained after the transformer's output passes through a dense layer with sigmoid activation. Binary cross-entropy loss is used to evaluate our model. Since the dataset contains almost 100 million rows of data, we simply use $97.5\%$ of data as the training data, and $2.5\%$ as the validation data. The ultimate hyper-parameters are as follows
\begin{itemize}
    \item Max sequence: 100
    \item Embedding size: 128
    \item Number of layers in encoder: $2$
    \item Number of layers in decoder: $2$
    \item Batch size: $256$
    \item Dropout:  $0.1$
    \item Epoch:  $10$
    \item Number of heads: $8$  
    \item Optimizer: AdamW
    \item Learning rate: $5 \times 10^{-4}$
\end{itemize}

\section{Results}
We did three experiments for comparison purposes: Transformer without lecture information, Transformer with lecture information, and LightGBM

\begin{table}[h!]
\centering
 \caption{Comparison of AUC Between Transformer Variants and LightGBM}
 \begin{tabular}{c c } 
 \hline
 Model & AUC  \\ [0.5ex] 
 \hline\hline
 Transformer with lecture & 0.800  \\ 
 Transformer without lecture & 0.800   \\
 LightGBM & 0.787   \\
 \hline
 \end{tabular}
 \label{tab}
\end{table}
\Cref{tab} shows that the multiple temporal granularities are significant such that even lecture information is ignored, they can capture the complex interactions between question difficulty and student performance. Moreover, LightGBM necessitated extensive feature engineering efforts. To avoid data leakage, which entails using future information to predict past information, each feature at a given time point in the model had to be computed based on its information up to that time point. Suppose we intend to employ the mean of lag time; in that case, we cannot just utilize the average time across all periods for each student. Instead, we need to consider the accumulated mean time up to time $t$ to circumvent data leakage. Another challenge when utilizing LightGBM is performing cross-validation for time series data. For the same reason, we opted for rolling window cross-validation. 

Therefore, our Transformer model has shown promising results in predicting student performance on the RIIID dataset, achieving state-of-the-art performance with fewer feature engineering efforts. This suggests that the Transformer's self-attention mechanism is able to effectively learn the relevant features from the data without requiring extensive manual feature engineering. Additionally, our model's ability to handle sequential data makes it well-suited for other time series prediction tasks in education and beyond. Overall, our proposed Transformer architecture offers a powerful and efficient approach for predicting student performance on educational assessments.
\section{Conclusion}

In this paper, we introduced a novel Transformer architecture for knowledge tracing, which is designed to capture multiple types of information that can affect student learning. Our architecture includes a sophisticated encoder component that captures information about the questions, parts of questions, and presence or absence of explanations, as well as a decoder component that incorporates temporal information about the sequence of interactions between the student and the learning materials. Notably, we included three separate embedding layers to represent lag time in different temporal granularities, allowing the model to capture the temporal dynamics of the data.

Our architecture also has the ability to model complex interactions between different factors that can affect student learning, including the relationship between a student's prior response and their subsequent performance on a related question, or the interaction between elapsed time and lag time in predicting student learning outcomes. Overall, our Transformer architecture provides a powerful tool for accurately predicting student performance and identifying areas where additional support or guidance may be needed.

\section*{Acknowledgment}
The author would like to thank Kaggle and Riiid AIEd for organizing the competition and providing such high quality data.  

\bibliographystyle{IEEEtranN}
\bibliography{reference}

\end{document}